# CONCEPTIVE ARTIFICIAL INTELLIGENCE: INSIGHTS FROM DESIGN THEORY

A. O. Kazakçı

## 1 Introduction – how can design research contribute to AI?

Artificial intelligence is a vast field of research whose output is shaping information and communication technologies that affect our economical and social life on a daily basis. On the rather rich list of topics relevant to research in artificial intelligence, it is surprising to note that creativity, a hallmark of human intelligence, is rather at the bottom of the list and not a priority in the mainstream research in AI. For a field whose ambition is to produce human level performance on tasks that require intelligence this absence is curious, to say the least. A major reason for this apparent lack of interest in studying creativity formally and implementing systems that aspire to be creative is the absence of clear, rigorous definitions of what creativity is. This is in contrast, for instance, to decision and learning, two processes for which more than half a century of efforts have led to precise formal frameworks enabling systematic and fruitful studies of these notions and building high-performance systems.

By contrast artificial intelligence, creativity has been studied intensively in design research, especially with empirical approaches. Starting with early work by Eastman (Eastman 1969, 1970), Akin (Akin 1978) and still others, a major approach that has been used in the study of designers thinking processes has been the protocol-based analysis. A large quantity of contributions has been produced (see e.g. (Cross 2001) for a review). One overall striking feature of a majority of this research is the quasi-total reliance on the problem-solving paradigm in the their interpretation of their results. There is now a growing consensus (see e.g. (Dorst 2006) or (Hatchuel 2002)) claiming that problem-solving (even in a broad sense of the notion) is too restrictive as a 'lens' to interpret design – which is a cause for concern in the interpretation of those results. Moreover, adopting the problem-solving paradigm as their conceptual framework for analysis, those studies do not contribute to the modeling of design – which is a cause for concern if design research is to find and develop its own authentic models.

In a paper called *natural intelligence of design*, Cross (1999) argues that design research should also contribute to artificial intelligence, and not only the other way around. The current paper adopts and expands on that position: Design involves possibly the richest forms of reasoning, providing thus a privileged context for the study of human cognition. We believe design reasoning is significantly different than ordinary reasoning situations since it involves the construction of previously inexistent objects. Rather than reducing design to other cognitive phenomena (such as problem-solving or incubation), design research should build richer models of design and creativity that would be useful for other disciplines as well. If we simply take formal models coming from AI and use them for describing design, we are restrained and forced to have design theories reduced to the reasoning paradigm underlying those approaches (i.e. decision and learning) –leaving out the possibility to study design creativity. By contrast, if we produce models specific to design, representing thus the specificity of design reasoning, we might be able to give back something in return to other disciplines.

## 1.1 Autonomous artificial systems requires conceptive intelligence

A most important chapter in current artificial intelligence research is to build systems that are autonomous. In most cases, if not all, what is meant by autonomous is the ability of an agent to behave within set limits in the way it is defined by the system-builder without the necessity of human intervention. This kind of design is enforced by an omni-present engineering concern: reliability of a system – which requires, in turn, predictable performances. Typically, on projects like Curiosity rover (Goth 2012), that entails significant investments, reliability of the system primes over most of other possible criteria. In such contexts, a system that has the ability of *surprising* its builders would rarely be a particularly desirable feature. For this reason, most of the efforts in AI have concentrated on planned and predictable behavior, independently of the technique used to implement intelligent functions. The hindsight is that, from this seemingly natural tendency of artificial intelligence philosophy and objectives, it can be seen that all incentive and aspiration for systems that can surprise their builders in a good way was also eliminated. This insistence on the construction of plans of actions (whether off-line or on-line), restricting voluntarily the domain, optimizing algorithms or programs for a specific, pre-defined and fixed set of tasks naturally prevented AI to look for systems capable of building new tasks in a creative way and that exploits old experiences for dealing with that novelty.

## 1.2 Beyond solving given tasks: Conceptive systems capable of designing original tasks

The present paper claims that truly autonomous systems require a specific form of reasoning that we call *conceptive intelligence*. By conceptive intelligence, we mean a capacity for an agent to design new concepts with respect to what it has observed (outside the scope of what is learned, e.g. by induction, over the observed objects) and to take necessary actions to build, realize or implement that concept. Such systems would thus be able to formulate new tasks continuously and try to solve them. There are numerous engineering and theoretical challenges for building such a system. These issues has been discussed in AI under the theme artificial life or open-ended evolution (Bedau 2003). The perspectives offered are based on traditional paradigms of AI, such as learning, interaction and randomness. In this work, we present an alternative view called imaginative constructivism, coming from design research (Kazakci 2013). Based on the creative reasoning process described by (Brouwer 1907, 1908, 1948; Heyting 1975; van Dalen 1981; Niekus 2010), this view suggests that design is a process by which the construction of objects proceeds towards conceivable and imagined properties. This is a dual constructivist process where creativity can occur both at the level of top-down generation of new definitions and the bottom-up generation of methods for building objects.

## 1.3 Brouwer machines: a model for conceptive systems

Given the above orientation, this paper defends the thesis that classical and foundational models in AI and related fields, such as decision and learning models, are implicitly based on some premises that we call *the-world-as-it-is* paradigm. We discuss basic formal models of decision and learning in order to explicate the differences of such models compared to design (section 2). This allows us to introduce an alternative worldview, namely *the-world-as-it-can-be*, based on the notion of design as imaginative constructivism (section 3). This framework allows us to discuss and analyze the traditional notion of search, omnipresent both in AI and design literature. We analyze a number of search processes and discuss their limits for describing design. In particular, we argue that combinatorial search can be seen as a construction process – *though it has hardly ever been applied in the dual constructivist perspective described by the notion of imaginative constructivism*. Building on those analyses, we sketch a model, called Brouwer machine, for a system that would incorporate a form of conceptive intelligence and discuss some issues related to its implementation. Last, we discuss the type of creativity that can be achieved by means of genetic algorithms. By interpreting this approach through our framework, we show that these are constructions machines rather then conceptive systems.

# 2 Revealing hidden limitations of traditional formalisms for conceptive reasoning: "the world as it is" as a paradigm

Traditional formal basis of models of decision and learning are implicitly based on a paradigm that analyses the world as it exists. In decision paradigm, decisions are taken about objects that exists or that are known to be feasible. In machine learning, the aim is to learn categories in a bottom-up fashion for objects that exists. Consequently, those formal approaches are not adapted for the creation of new objects. Let us discuss the properties of those models based on the underlying formalisms.

**2.1 Decision paradigm as evaluation of known objects**

Tsoukiàs (2008) defines a generic evaluation model into which a large part of the existing decision aiding models and methods can be fit. His model is an n-tuple:
$M = <A,D,E,H,U,R>$, where
- A is a set of objects (alternatives, solutions) to which the model will apply;
- D is a set of dimensions (attributes) under which the elements of A are observed, measured,
- E is a set of measurement scales associated to each element of D;
- H is a set of criteria under which each element of A is evaluated;
- U is a set of uncertainty measures associated to D and/or H;
- R is a set of operators enabling to obtain synthetic information about

the elements of A or of $A \times A$, namely aggregation operators (acting on preferences, measures, uncertainties, etc.).

The distribution of the various parameters in the model irrevocably denunciates where the majority of the efforts have been concentrated in the decision literature. With the exception of A, all the elements of the model are destined to measure and compare the properties of the alternatives. Objects from A are described on some scale along different dimensions. Whether there are some uncertainty measures or not, an aggregation procedure compiles this various information, where usually some information is lost, but an overall evaluation is reached at.

It is claimed by Tsoukiàs (2008) that this model can accommodate almost all major formal decision techniques and approaches. It is instructive to note that this model can only function if all the model parameters are supplied, or else, there can be no evaluation. Compared to decision processes, in design situations most of these information do not exist and cannot be collected simply by asking participants to the process. In particular, the set A is empty prior to the process (Hatchuel and Weil, 2002). It is precisely the aim of the design process to construct the objects that are called alternatives. Innovative design is such an important process in the current economic processes because we do not have alternatives to many challenging problems. Most of the writings about decision paradigm considers the inability to take (correct) decisions will lead to a crisis. Yet, major crises occurs when no alternatives exists when an action is necessary.

It should be remarked that the decision paradigm does not consider the question of generating objects. Albeit there exist some work, mainly in engineering design literature, using the terms 'generation of alternatives' by means of evolutionary computation. In those works, what is being generated are a discrete and finite set of alternatives, typically on the Pareto frontier given by some set of criteria, from among an infinity of solutions (called, the feasible solution set) that are assumed to already exist and feasible. Thus, no new object is really being created and object that are unfeasible at the beginning of the process are not even considered.

In decision and evaluation models, objects being defined at the beginning of a process and assumed to be feasible, the only knowledge that is being derived from the process is the preferential information from what is already known about the objects. On the other hand, design focus on the construction of new objects:

Another way to characterize the difference between decision and design processes is to consider the notion of "states of nature" omnipresent in the decision models. In decision, the main source of uncertainty and ambiguity is often conceptualized as an uncertainty measure (e.g. a probability distribution or a possibility relation) over the states of the natures. Most approaches dealing with such structure either try to reduce the uncertainty or to take decisions in such a way that it will optimize some estimated outcome (e.g. minmax regret). Note that it is never the question of changing the world to provoke the creation of new states of nature: it is considered that the world can be changed by means of the actions of an agent (thus, moving forward to some alternative state of nature) but no new worlds are created. Design models should go beyond this framework since what is of interest is the passage from one system (states and transitions) to the next – and hopefully towards a more fruitful one. A related question that we shall try to shed some light later on is "how does an artificial program provoke new states of nature?"

(Bouyssou *et al*. 2013) argue that evaluation tools "are a consequence of the decision aiding process" and they should not be chosen before the problem has been formulated or the evaluation model constructed. On the basis of the discussion so far, we argue that evaluation models (and tools) are a consequence of a design process, especially in the case where there is no feasible alternatives at the beginning of the process and not the other way around.

A simple model that explains the evaluation process would be

$$X, f \rightarrow Y$$

where X is the set of alternatives, Y= f(X) is the evaluation and the task is to determine D,E,H,U and R to build f and apply it on X. No new objects are created; at best information about already feasible alternatives are discovered and compiled by means of formal processes to reach additional information, called evaluation. Evaluation models operate indeed within a world-as-it-is paradigm, without changing the world by means of creation of new types of objects.

## 2.2 Learning algorithms as eliciting consequences of known objects

Machine learning approaches have also been designed to operate on known objects. Two major paradigms are inductive and deductive learning. In deductive learning approaches some database of knowledge is used to produce new knowledge under some given closure operator. For example in propositional logic, modus ponens allow discovering the consequences of known facts. Given a theory T = (P, P → Q) and an operator of deduction ⊢, we can derive T ⊢ Q. Levesque calls T explicit knowledge, where as Q, the logical consequence of T, is implicit knowledge (Levesque 1984). In such a learning mechanisms, any notion of novelty would be deceptive. When Q is revealed through deduction, some new facts are indeed learned but no new objects have been created. At the very best, in more general cases such as first-order logic deduction, new knowledge about already existing objects will be produced.

In inductive learning approaches, the aim is to extrapolate relationships from a set of observed objects, called a training set, so that accurate predictions about future examples can be made. More formally:
- $(x_i)_{i=1..n}$ are observations (objects) where for all i, $x_i \in X \subseteq R^p$;
- $X = (x_{i,j})_{i=1..n, j=1..p}$, the matrix of observations;
- $Y = (y_i)_{i=1..n}$, the observed output;

In a classification problem the output corresponds to C classes; where each individual object needs to be assigned (e.g. good, medium, bad grades for students). In a linear regression problem, the output will be a continuous variable; $Y \in R^n$.

Independent from the specific algorithm used or the number of the output classes, learning problems formulated in this fashion aim to build a function f representing as much as possible the mapping that exist between the observed objects and the corresponding outputs. More formally, assuming that $(x_i, y_i)_{i=1..n}$ are realizations of some random variables $(X_i, Y_i)_{i=1..n}$ from some unknown distribution over existing objects, the aim of the learning algorithm is to approximate a function f such that $Y^- = f(X)$ that is close to Y given some distance metric l. The function f is called a predictor and the metric l is called a loss function. Much of the machine learning literature revolves around this notion of loss function: algorithms are optimized for specific cases of learning problems in order to handle problems related to loss functions, accuracy of the predictions for future observations in terms of recall and precision. A basic model that explains this approximative (or predictive) stance is thus;

$$X, Y \to f$$

where X, Y and f are as defined above. With respect to design, we can see that what is created here is not new objects, nor new classes of objects (the outputs). It is a function that is hopefully a good representation of the mapping between objects and their corresponding classes. What is sought is not the creation of new classes of objects, to for which corresponding objects are built, but a search for a fit between the available and already existing objects and their known categories. Learning models are not designed to create new objects nor to use existing available data to hypothesize about the possibility of objects that may not even be contained in the available data. They do not consider what cannot be found in the current data but that could be interesting to know. They neglect thus the fact that sometimes it is simply more important to decide what to look for then finding what is already there. As such, although different from the evaluation models, learning algorithms also operates assuming implicitly that the world exists as it is and by applying a series of operations we can learn new knowledge from that world by generalizing upon existing objects or relationships.

## 3 Design reasoning as "the world as it can be"

### 3.1 Design as the evolution of object definitions

By contrast to decision and learning paradigms, design is the creation of some new object. Design theories and models try to capture in various ways this basic and fundamental observation. Most theories of design could be described, one way or another, as the evolution of some description D of the designed artifact (i.e. $D_1 \to D_2 \to ... \to D_m$). For instance, in Schön and Wiggins's description (Schön and Wiggins 1992) a transition from one description to another occurs with what is called a 'design move'. In topological spaces (Braha and Reich 2003), the descriptions are 2-uples of the form $<F_i, D_j>$ where transitions may either change $F_i$ (e.g. functional descriptions) or $D_j$ (e.g. structural descriptions) thanks to operators called 'closure' (e.g. deductive closure). A majority of such design models, if not all, do not attempt to describe the creative mechanism leading such transitions. An exception is C-K design theory (Hatchuel and Weil, 2003). Compared to most of other formal design process models, C-K theory describes the evolution of design definitions by placing knowledge at the heart of the transitions. Said in other terms, knowledge is a necessary resource for generation new descriptions: Objects descriptions are created and evolve using knowledge, taken in the form of logical propositions with a truth value.

### 3.2 Design as imaginative constructivism

Going further (Kazakci 2013) brings some new perspectives concerning the elaboration of definitions in design. He argues that constructivism is a foundational issue in design research and he studies forms of constructivism in the elaboration of object definitions in design. Several such forms exist in design literature. First, we can find a social constructivist approach, for instance in (Bucciarelli 1988), where object definitions are constructed collectively over time. Second, we can identify an interactive constructivism, for instance in the work of (Schön and Wiggins 1992), where a designer interacts with some media to progressively construct a design. Aside from these traditional forms, Kazakci contends there is a third form, called imaginative constructivism. Imaginative constructivism pleads for a world view that, in innovative design, new *types* of objects are imagined while methods for building or implementing those objects are sought. This is a dual constructivist process, where the construction of the object definition interacts with the construction of the method that would allow building or implementing that object. On both of these processes, i.e., in the construction of the type or the method, it is possible – and often, it is required, to introduce novelty.

Foundations for the notion of imaginative constructivism come from the study of a particular design domain – the mathematics. Kazakci (2013) studies Brouwer's Intuitionism, one of the major constructivist approaches to mathematics that captures several fundamental properties of design reasoning. First, it explains mathematical activity as a reasoning process performed over time. Second, it puts emphasis on the *constructability* of objects, rather than the truth of their existence. Third, it acknowledges the incompleteness of knowledge and the possibility of constructing new objects. Fourth, the construction of unprecedented and unpredictable objects is taken into account by a notion of creativity of the mathematician: her *free choices*. From a design theory standpoint, this is a significant feature of Intuitionist mathematics. Allowing an act of free choice at any moment and the possibility to break away from any fully determined (lawlike) object allows the consideration of *partially determined objects* with *novel properties*. This conception recognizes the creative nature of the mathematical activity. There is always the possibility to continue defining an object in a way that distinguishes it from all the others that are known so far, creating thus a novel object (van Dalen 2005).

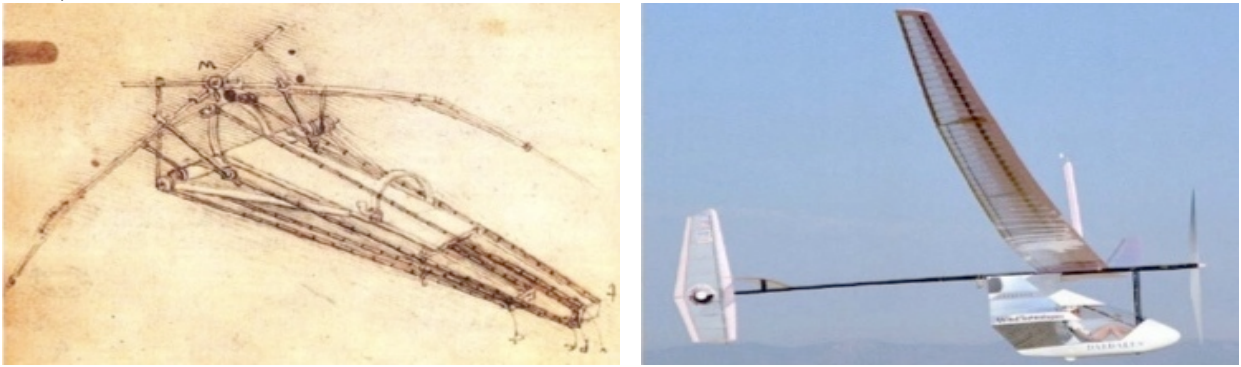

**Figure 1. On the left, Leonardo da Vinci's human flying machine concept. On the right, the Daedalus human flying machine. Da Vinci's *imagined what* a human powered flying machine would be like centuries before a working prototype was *built* – the Daedalus – by a group of NASA researchers and MIT engineers (Eris 2006).**

**3.3 The clash of imagination vs. knowledge**

Kazakci (2010) defines imagination as the ability of the mind for creating thoughts that are not being (or that has never been) experienced (such as a horse with wings, a flying chair, a mobile phone preventing heart attacks). It is suggested that this definition is fundamental to understand human creativity. Strikingly, it has been ignored, or at least, not directly accounted for, in models of creativity. On the other hand, human beings are programmed to be sense-making machines creating a meaning for their experience (Bartlett 1932; Clancey 1999). Throughout our education and everyday experiences we are often required to reject, ignore, dismiss implausible or absurd ideas (*normally*, chairs do not fly, horses do not have wings and phones do not prevent heart attacks).

Based on these premises, we can see the difficulty in the creation and evaluation of fundamentally new ideas: our tendency to look for meaning and possibility may prevent us to push our imagination towards new ideas. New ideas may seem, by their very nature, strange, or meaningless preventing the designer to recognize any value or investigate further. Nevertheless, it is the capacity of the designer that allows her to look past the domain of known objects with known properties, to formulate interesting combinations of properties for objects yet-to-exist and start genuinely innovative design processes where new classes of objects can be bred.

### 3.2 The "what" and the "how" of an object definition

Mathematics is a particular design domain where new objects with interesting properties are sought (Kazakci and Hatchuel, 2009; Kazakci, 2013). A fundamental issue that causes strong oppositions among mathematicians concerns directly design theory: the definability of objects. Constructivist mathematics defend the argument that an object can only be defined and be made to exist if it can be constructed explicitly by a method. This view on existence of objects can be called *existence as constructability*. A method is a set of ordered operations that transforms an input to an output. In this sense, we can use equivocally a method, an algorithm or a proof. By contrast, non-constructivist mathematics accepts object definitions whose existence can be proved by logical means (i.e. without a method or an algorithm of construction, proven solely on the basis of given axioms by means of logical deduction). This view on existence of objects can be called *existence as truth*. This distinction gives us the fundamental dichotomy between constructivism and non-constructivism: Either, the world exists and it should be studied as such. Or, the only way to guarantee the existence of the world is to construct the said-world (nothing that we cannot construct exists).

In design, these two extreme positions are overly strong and they may easily become false depending on the context. In the world, there are a variety of things that already exist about which some particular actor might know or observe some properties without knowing how to construct or reproduce an object with those properties. They are nevertheless useful and we can make use of them, choose them over other objects or learn about their properties without knowing a method to construct them (be it individually or as the society). Thus, both modes of thinking are used: *what* the object will be and *how* the object will be built is constructed together and interdependently (Kazakci 2010).

Such dynamics are reminiscent of the distinction between the formulation of a theorem and its demonstration by a proof. It is known in mathematics that, in some cases such as Fermat's last theorem, there have been several centuries between the two processes. In design literature, it is possible to find cases that highlight similar dynamics. For instance, Eris (2006) describes an example of the human flying machine of Leonardo da Vinci that inspired centuries later Daedalus built by NASA engineers; Figure 1. Examples such as this one are indicative that there may be various processes of construction in design processes (e.g., the construction of a definition vs. construction of an actual object). Recent experimental data (Eris 2004 ; Edelman 2012) also supports the idea that designers think and act differently when thinking about *what the object should be* or *how the object can be built* (Kazakci 2010). The imaginative constructivist dynamics allow thus to reveal a dual constructivism in design processes. This issue has been under-investigated in design literature, often collapsing both notions into a single one. Although theories of co-evolution (e.g. problems-solutions, concepts-knowledge, functions-structures) exist in design literature (Maimon and Braha 1996; Braha and Reich 2003; Hatchuel and Weil 2009), either they do not explicit the constructive aspects or they do not take into account the free-choices of the designer.

## 4 Collapse to mono-space search: implicit elimination of a duality in AI

Given the world-as-it-can-be paradigm described in the previous section, we shall analyze in this section some particular techniques from the AI literature in order to better understand their relationship with design. A fundamental notion in traditional AI is *search*. When the cardinality of the set of solutions to be considered is huge or even potentially infinite, search procedures are used to find solutions with desirable properties. While the notion of search is omnipresent in AI and it has been generally accepted as being inevitable, it has been rarely discussed what the search *creates*. It has been made abstraction of what kind of objects are manipulated during search and why, and the focus remained rather on optimizing search performance (i.e. search time). We shall discuss two very common examples of search in order to discuss and distinguish two very contrasted purposes that usually goes unnoticed. Let us remind that search related models (i.e. problem-solving) has been considered for long time as an adequate model for design (see e.g. (Cross 2001) for examples and (Dorst 2006) for a discussion).

## 4.1 Linear programming search: search without construction

An archetype of search problems is search for optimal solutions to mathematical linear programming problems. In the case of maximization, a general form for a linear program is the following:

$$\text{Max } \mathbf{z}.\mathbf{x}, \text{ s.t. } \mathbf{A}.\mathbf{x} \leq \mathbf{b}, \mathbf{x} \geq \mathbf{0}$$

where $x = (x_1,...,x_n)$ are the real-valued variables and $z = (z_1,...,z_n)$ are unit profits that defines the (economic) objective of search. The constants $\mathbf{A}$, $\mathbf{b}$ and $\mathbf{0}$ circumscribe an acceptable domain of variation for the variables. The aim of the search algorithm is to find, among all feasible objects respecting the constraints, a particular one that optimizes the economic function (the objective).

In this formal problem, we do not have an explicit list of the individual objects. Rather, we have knowledge about some properties they all satisfy (the constraints that gives what is called a feasible region) and a means for measuring their quality (the objective). Let us remark that, given the form of the feasible region, a solution always exists and all the represented objects are supposed to be feasible. Several search algorithms that guarantee an optimal solution are known (e.g. simplex method or interior point algorithms).

We do not know in advance what the optimal solution will be, but we know for sure what type of solution it will be: This is already characterized by the constraints and even more so by the dimensions $D = (d_1, \ldots d_n)$ on which all of the objects considered in the problem formulation has been defined once and for all, before the algorithm is applied. The 'what' of the objects have been determined and the aim of the algorithm is not to explore alternative definition types. Said in other terms, dimensions or constraints do not change during the search. Compared to a design process as discussed in previous section, neither a new definition is created, nor, an object fitting that definition is constructed.

Let us also note that in traditional sensitivity analysis or robustness analysis those dimensions D do not change either. Hence, any variability in search parameters do not provoke a new type, only a different optimal solution. Aside from the stability of object definitions, another major point of interest for us is that the optimal solution is guaranteed to be from among the feasible solutions. The purpose of the search is not to seek for solutions that may be unfeasible for the moment but interesting nonetheless. At least, in this context, search is not aiming at finding imaginary (or, fictive) solutions that are yet to be made feasible.

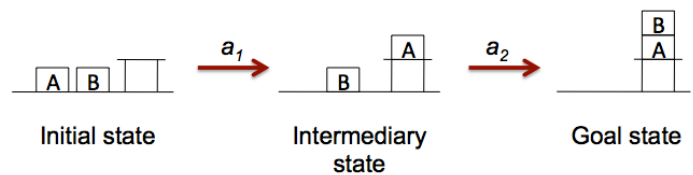

**Figure 2. A simple blockworld diagram. The initial state is transformed to the goal state by means of application of actions $a_1$ then $a_2$.**

## 4.2 Combinatorial search: search with construction…of?

Search formalisms has been extensively applied in planning tasks as well (Fikes and Nisson 1971; Bonet and Geffner 2001). To get a better understanding of what kind of objects are elaborated during AI planning, we shall consider one of the most conventional examples used in the development of planning programs, namely, the block worlds (see Figure 2) (Nagata *et al*. 1973; Russell and Norvig 1995). Blockworld is an abstract, closed world for experimenting with AI systems. This conceptualization was particularly useful for building early systems for planning and robotic navigation. It consists of a flat surface, a set of blocks and a robot (or a robotic arm) able to move the blocks around by applying some actions (e.g. pickup() or stack()). States are descriptions of the world by means of predicates (e.g. on(A, B) or onTable(A)).

Starting with an initial configuration (or, state), the program has to search a combination of actions to reach a final, desired configuration, called *goal state*. Often, there is no trivial solution for combinatorial search and the program might get stuck (e.g. in the above example, if B is put on top of the table before A) in which case backtracking is necessary. The search starts over from a previously explored configuration that is evaluated to be likely to lead to a solution. The program stops if the available computational resources are exceeded or a sequence of actions leading to the goal state has been found.

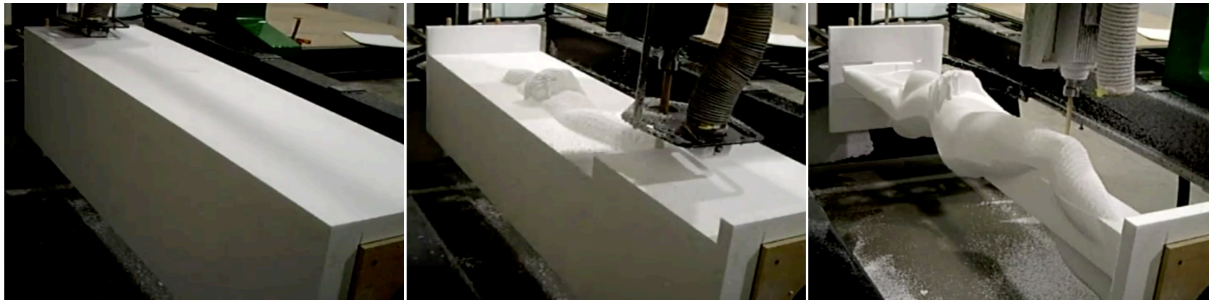

**Figure 3. The transformation of some input material by a CNC machine via a set of planned actions.**

This model has been used extensively as a valid model for studying human reasoning skills in cognitive psychology and in design research (Cross, 2001). In a design context, the goal state has been interpreted as the design to be reached at, i.e. a set of requirements. A designer needs to take some actions to manipulate objects (or, their representations, i.e. sketches) about her to reach the goal state. This view of design raises a set of legitimate questions (see e.g. (Dorst 2006). In our context, the main question is what kind of object definitions are being created and manipulated during reasoning?

There are two things that are created by this process. First, a plan of actions – with a clearly defined input and output set – is created. In terms of the discussion of previous section, this corresponds to a *method* of construction (or, to a proof, for that matter (Kautz and Selman 1998)). The search creates a set of instructions (as an algorithm) that *transforms* some input to some output. In design and manufacturing related fields, a standard application of this type of search is the use of early CNC machines; see figure 3. Second, the output configuration that corresponds to the goal state will be created by the execution of the plan. In the example of figure 3, this is a sculpture of a female body. In the most general case, this would be a re-arrangement of the initial input objects.

Where is design? The answers to this question were ambiguous in the literature precisely because the interpretative model used to define what design is the same as the model that is being interpreted! Our overview of the model of combinatorial search based on our imaginative constructivism framework sheds light on what is really being designed (and to what extent). The entity that appears and that allows the creation of the same artifact as many times as it is applied is a method of production. The 'sculpture' in this example is not the design; it is the first of the many output objects that can now be produced. The design is the method. Thus, when combinatorial search is applied to a set of actions, what is being designed is a constructive proof of existence producing a first example of the new type of object (assuming the system has not produced a plan for the same goal state before).

Hence, the interesting question for us is: where does the *type* (definition) come from? Is it new? The answer to this question is now straightforward: the design of the 'type' of new object to be constructed has finished before the combinatorial search process has started. It was the goal state given to the program as a parameter. Whether it was new or not cannot be determined based solely on the input-output pair of the search process. We need to be able to define a reference library of types that is specific to each designer that contains the old objects and their types in order to be able to determine if a particular type of sculpture is new. Therefore, there is no creative design of types, nor creative design of a method (only, known actions are combined) – only a novel sequencing of actions in order to create a new method for constructing an object with a stable definition (of what that object is).

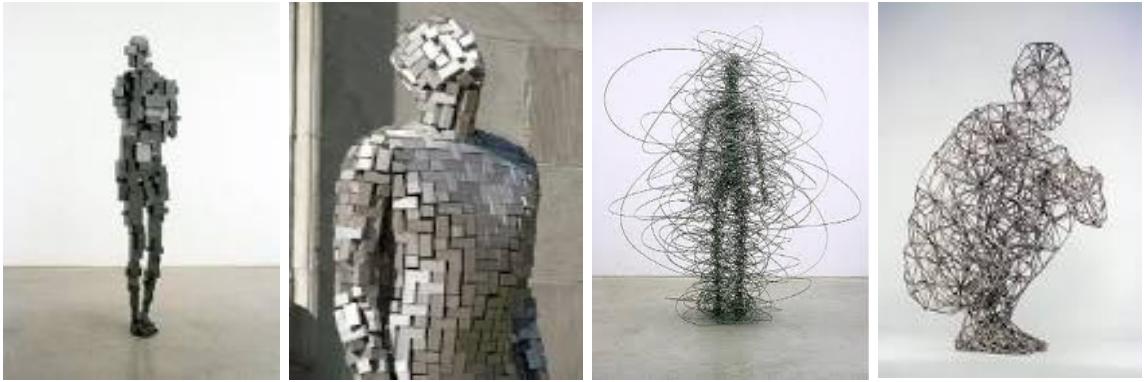

**Figure 4. Different definitions of 'human body' – the creation of such 'types' are not taken into account by standard combinatorial search over action sequences**

Unless the search process can change the type definition of objects on the run, there is no possibility to imagine new types of objects (in this setting, new goal states). In the previous example of body sculptures, new goal states are simply new representations of the human body; Figure 4. In traditional combinatorial search this aspect does not exist, since the combination is only applied to individual *actions* to build methods. As we have seen in section 3, the imaginative constructivist nature of design presupposes that the creative design act requires the combinative creation of both type descriptions and methods. In AI programs one or the other of these two dimensions are not treated. *Search is applied only on one dimension of the object definition, collapsing thus the creative dynamics* we described previously to a mono-space search. This in unfortunate insofar as the real richness of a design reasoning occurs at the interaction of these two types of definition construction processes (Kazakci 2013)

Another shortcoming of this type of search process, in terms of its adequacy for incorporating conceptive intelligence, is its inability to creating goals – that might even be not achievable by the current set of actions. This assumes not the ability to formulate goal states, but also goal states that cannot be reached at by the current set of actions. In such a case new actions need to be learned. Systems able to learn new types of actions, thus enabling the realization of new types of tasks, exist in AI and robotics literature (see e.g. (Konidaris *et al*. 2010)). However, a dual constructive search mechanism, allowing the system to build new and unprecedented tasks, while acquiring new types of actions by means of interaction with the environment do not exist. This, however, is the *sine aqua non* of a conceptive intelligence for autonomous creativity.

## 5 Brouwer machine: a conceptual model for systems with conceptive intelligence

According to the imaginative constructivism framework, design implies a dual constructivism on the definition of types of objects and the methods by which they are produced. This type of process involves the articulation of top-down and bottom-up processes. Both the construction of definitions and the construction of methods may be changed significantly during the activity by the free choices of the designer. Bases on these premises, the first feature of the model we introduce is a language with two components.

### 5.1 Dimensions of definition

The overall idea behind our notion of Brouwer machine is that there are two fundamental dimensions through which a class of objects might be specified:
- The type definition – Part of the definition of an object stating *what* the object is through the specification of its properties
- The method definition – Part of the definition of an object stating *how* the object can be built through the specification of a sequence of actions

Let us have a set of properties $P^T = (p_1, p_2, \ldots, p_n)$ accessible to the system at time T. Let $L_P^T = (c_1, c_2, \ldots, c_N)$ be the library of conceptual descriptions at time T. Each $c_i$ is a non-empty subset from $P^T$; i.e. $\forall i, c_i \in [(\Pi P^T)\backslash \emptyset]$ where $\Pi X$ stands for partitions of set X.

Let us have a set of actions $A^T = (a_1, a_2, \ldots, a_m)$ available to the system (i.e. through its effectors) at time T. Let $L_A^T = (\pi_1, \pi_2, \ldots, \pi_M)$ be the library of procedural descriptions at time T. Each $\pi_k$ is a non-empty subset from $A_T$; i.e. $\forall k, \pi_k \in [(\Pi A^T)\backslash \emptyset]$. Note that $L_P$ allows defining goal states whereas $L_A$ allows defining plans of action for reaching those goal states.

### 5.2 Free choices by novelty-driven search on definitions

The aim of the system is to mimic a kind of design reasoning by continually alternating between:
- the generation new type descriptions $c_{new}$ such that $c_{new} \in [(\Pi P^T)\backslash \emptyset]$ and $c_{new} \notin L_P^T$
- the generation of new plans of actions $\pi_{new}$ such that $\pi_{new} \in [(\Pi A^T)\backslash \emptyset]$ and $\pi_{new} \notin L_A^T$ and allowing the construction of $c_{new}$

There are several points to be considered. First, the system needs a mechanism to find new $c_{new}$ and $\pi_{new}$. One way of generating these entities is to novelty driven search (NDS). This might for instance be a genetic algorithm pushing for novelty rather than fitness with respect to predetermined criteria (Lehman and Stanley 2008). In each iteration, privileged solutions are only those that are newest (i.e. farther apart) from the existing set of solutions. Taking advantage of the crossover and mutation operators genetic algorithms are indeed able to build a myriad of combinations of existing entities, some of which will not exist in the current libraries. Other solutions are envisageable but will not be discussed in the current work.

Second, for any $c_{new}$ generated, it is likely that the system will not have an existing $\pi_k$ able to build $c_{new}$. Thus, the imagination of a new enunciation will trigger the necessity to imagine new plans of actions. Let us also note there is an injective mapping from $P^T$ and $A^T$, in the sense that for each $c_i$ there may be multiple plans $\pi_k$ building it. The inverse is not true.

Third, the system will not be able to generate new plans and concepts indefinitely. The combinatorial search among $L_P$ and $L_A$ will eventually exhaust all possible new combinations – at which point, no further imagination is possible for the system (with respect to the definition of section 3). To prevent this and in order to progress towards truly autonomous agents, the system needs another operator that allows it to communicate with its environment and to be able to add new elements to P or to A to enrich its design languages.

### 5.3 Example and discussion: Mazes and escape artists

An example Brouwer machine might be set up in a hypothetical domain consisting in the design of mazes. Mazes are typical examples for traditional search programs in AI. In our case, rather than having a solver for a particular maze, a Brouwer machine would be an escape artist, not only solving mazes but aspiring to create new ones in a continuous manner – such that every new maze would be more interesting to solve in some sense (i.e. more challenging), with respect to the currently known mazes. In such an application, the system would generate mazes, always targeting newer mazes that are different. For each maze, it would search a plan of escape (e.g. evolving plans of actions until a plan that solves the maze is found). At first sight, this might seem as a standard co-evolutionary process. There are several significant differences. First, we are interested in finding a repertoire of mazes and escape plans, rather than individuals overcoming each other by mutual co-evolution. It is not the maze versus the escape artist that is being evolved. It is a designing entity, the Brouwer machine, which explores the creation of new mazes and plans for solving them quickly. Contrary to typical co-evolution logic, a newly generated maze will not be a maze that the plan from the previous iteration would not be able to solve (hence, not necessarily a better maze, according to some criteria). Instead, it would simply be a newer one – with respect to the previously explored mazes. Each plan will solve only one maze: it will be a method for that instance only.

Even with this simple example, some limits for implementation already appear. First, the maze domain is too limited in the vocabulary it allows for $L_P$ and $L_A$. For instance, whatever the new maze the system imagine, it suffices to use the same set of navigation actions to solve it. In this example domain, there will be no opportunity to extend $L_P$ and $L_A$ eventually. In contrast, if we consider a Brouwer machine set up on a digital image or physical prototyping domain, the scope is considerably broadened since the possibilities are endless.

Second, the iterative alternation between the maze definitions and solver definitions are rather simplistic. More interesting would be some mechanism by which an interaction with previous mazes and their solvers can be achieved to foster the dual construction process. There are at least two types of interactions we can consider:
- How old solutions can help a particular maze?
- How old solutions and mazes can help produce new mazes (e.g. different and showing particularly interesting behaviour, such as increased level of difficulty without any explicit objectives).

In realistic design situations, those interactions immediately become complex. They are not taken into account by the current framework, but they constitute a necessary step to progress towards conceptive intelligence.

### 5.4 Central problem of conceptive intelligence

The previous remarks allow us to finally state and discuss what we deem as the central problem of conceptive intelligence. In what we called an imaginative constructivist process, a dual constructivism with free choices can occur on both properties and methods defining objects. The central ability for a Brouwer machine, or any designer thereof, is the ability to choose which novelty will be pursued and elaborated. As mentioned, in a realistic design setting, an expert designer would generate more than one novel object definitions that might be explored next. The true mark of an expert is to better judge which of the currently considered novelties is worth exploring given the available resources. Said in other terms, it would be a decision mechanism not for choosing a best among existing objects, but a most interesting to explore among a set of novel definitions. Work initiated in (Hendriks and Kazakci 2011) offers some perspectives on the logic of this issue. As signalled in (Kazakci 2013), this is a decision theory specific to design processes – that is yet to be formulated.

## 6 Discussion: Are genetic algorithms creative?

### 6.1 Genetic algorithms as a mean for scientific discovery

Genetic algorithms offer a powerful approach for combinatorial search. Based on the metaphor of natural evolution, a genetic algorithm maintains a population of candidate solutions for the problem at hand, and makes this population evolve by iteratively applying a set of stochastic operators. At each iteration, a subset of the population survives and is given the opportunity produce offsprings. The survival of candidate objects depends on some evaluation criteria, called fitness. Genetic algorithms have been often associated with creative processes (Koza 1999; Koza *et al.* 1999; Renner and Ekárt 2003) since, given a problem formulation, the solution space can be explored conveniently by evolving candidate solutions in various directions.

A recent application of genetic algorithms that offers promising perspectives is in the field of discovery of scientific laws from experimental data. (Schmidt and Lipson 2009) have been able to *distil* free-form natural laws from motion-tracking data captured from various dynamic physical systems ranging from double pendulum to harmonic oscillators using genetic algorithms. Starting with symbolic expressions (e.g. +, /, sin(), ω, θ, etc.), the algorithm was able to generate more and more complex sentences by combination. The representation of a symbolic equation in computer memory is a list of successive mathematical operations. The construction of such symbolic expressions that fits a given dataset is a traditional application in the field of data-mining called symbolic regression.

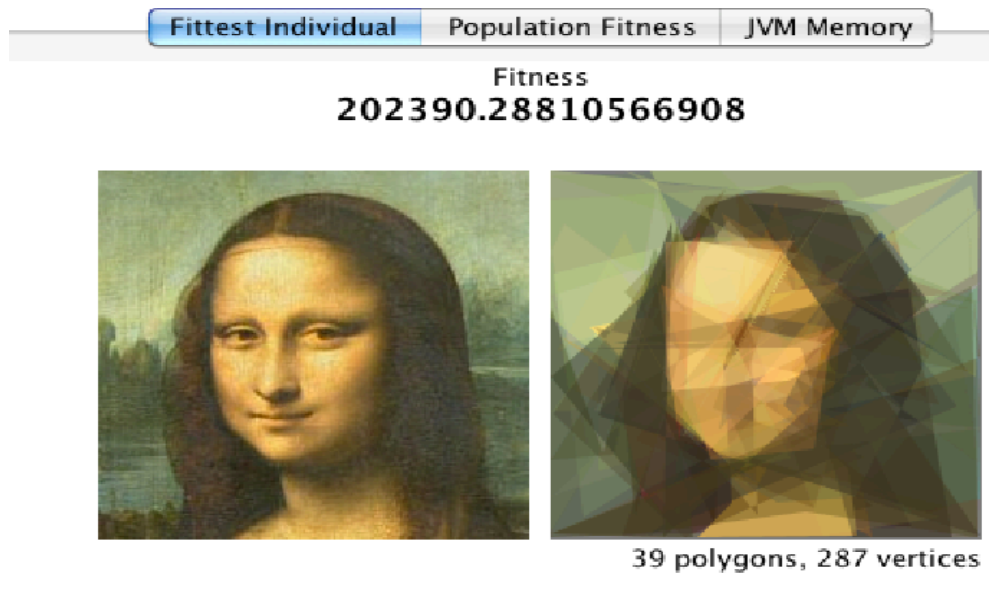

**Figure 5. On the left, the picture of Mona Lisa as a 'type' representation; On the right, 'an instance' approximating the type using a combination of polygons – pictures produced using Watchmaker framework**

The particular insight Schmidt and Lipson propose is a principle for identifying non-triviality. This insight is based on the observation that, in order to claim a fit, between the generated symbolic expression and the data observed, the partial derivatives of both the symbolic expression and from the numerical data should vary in the same way. Using this principle as the fitness measure and the genetic algorithm over symbolic expressions they were able to generate complex invariants such as Hamiltonians, Langrangians and equations of motion for system of various complexity. The type of law discovered depends on the type of variables provided to the system on a given run. Symbolic expressions obtained from simpler systems have been found to be effective in bootstrapping search for more complex systems.

Without prior knowledge about physics, kinematics, or geometry, (Schmidt and Lipson 2009)'s system detected complex relationships such as nonlinear energy conservation laws, Newtonian force laws, geometric invariants, and system manifolds in various synthetic and physically implemented systems. And it is claimed that many applications exist for this approach, in fields ranging from systems biology to cosmology, where theoretical gaps exist despite abundance in data and that scientists may use tools such as theirs to focus on interesting phenomena more rapidly and to interpret their meaning.

*6.1.1 The intelligence of genetic algorithms – conceptive or not?*

Schmidt and Lipson's system is a powerful discovery engine. Since discovery is a notion that we can relate to intelligence, the question that is interesting for us: is it intelligent? Is it creative – Considering the fact that it discovers complicated laws of nature without any notions of physics or domain knowledge? There is indeed a tendency to assume that genetic algorithms are, in a sense, intelligent. For instance, (Reynès 2007) considers that genetic algorithms incorporate intelligence at the level of selection of survivors – and that, this is the only intelligent step. Said in other terms, it is claimed that the intelligence of the genetic algorithms is in the selection mechanism. Jacques Monod, Nobel Laureate in Biology defends this very view: « Many distinguished minds appear not to be able to accept, nor to comprehend that, from a source of noise, the *selection* was able to, by itself alone, pull all the music in the biosphere. »

The selection step in a genetic algorithm is based on a metric representing the membership (or the distance from) a type definition; Figure 5. As we have stressed previously, creating a type and building an instance of that type are two separate things. As can be seen from the Mona Lisa example in Figure 5, the only thing selections does is to evaluate how far away an instance is from the type defining its category. What does the genetic algorithm in this context is simply to *construct* an instance of a type – given the type. If any intelligent act occurred in this context, it is not during the selection, but during the type creation – by Da Vinci himself, in this example. Contrary to what Monod claims, it's not the selection that pulls the music from a continuum of noise - It is a design effort that determined, before even the process of construction has started, *what is a nice noise*.

*6.1.2 Interpreting Schmidt and Lipson's system through Brouwer machines*

Now that we have qualified genetic algorithms as a construction machine (just like the combinatorial search process discussed in section 4.2), we can discuss Schmidt and Lipson's system with respect to the notion of Brouwer machine. A point that matches immediately is the symbolic expressions of equations. Those can be easily mistaken as types. Howeber, when we consider that they are stored in memory as a list of successive operations, it is easier to see that this list of operations can be seen as plans of actions, which, when applied, reproduces an instance similar to what has been observed. $A_T$ are all the elementary operations and constants (e.g. $\omega$, $\theta$, $\sin()$, $+$, etc). $L_A^T$ is then all the sentences that have been formulated by combination at iteration T (e.g. $\sin^2(x_1) + \cos^2(x_1)$ or $x_1 + 4.56 - x_2 x_1/x_2$). Note that the instances (streams of a motion data captured by a camera) are not perfectly generalized by the system. Thus, the application of a plan discovered along the way will not produce the signal that was observed either. Note also that the methods formulated by the system are not created *per se*. They are generalization over observations using a convenient construction machine. This also implies that the system will not have any free-choice on the methods created – since it does not target any act of creativity or even intelligence: it only aims reproducing accurately what has been observed.

Concerning the set $P_T$ we can see that there is no properties, or any combination thereof, that are considered by the system. Consequently, there is neither type creation, nor any attempt to create new types by free-choices. That would have been a whole different story, had the system the ability to think of new types of natural laws – and then take action and provoke a change in the world to test its set of methods. This takes us back to the central question of conceptive intelligence: how would the system know what novel type to create and to try to build? We see that Schmidt and Lipson's system, despite its power and accuracy for the task domain with respect to which it was built, is not a system with which we can hope to create an autonomous explorer that would conceive any new scientific concept. In its current form, rather than distilling laws, Schmidt and Lipson's system distil only regularities within the observed data – that would be recognized as theoretical concepts in physics by an expert on the matter, already familiar with that concept. Without such knowledge and the ability to formulate preferences on unknown types obtained by free-choices, it is not possible to have systems with conceptive intelligence.

# 7 Conclusion

The current paper offered a perspective on what we termed conceptive intelligence – the capacity of an agent to continuously think of new object definitions (tasks, problems, physical systems, etc.) and to look for methods to realize them. This framework we call Brouwer machine is inspired by research in the design theory and modelling, with its roots in the constructivist mathematics of Intuitionism. The dual constructivist perspective we described offers the possibility to create novelty both on the types of objects and methods for constructing objects. More generally, the theoretical work on which Brouwer machines are based is called imaginative constructivism. Based on the framework and the theory, we discussed a number of paradigms and techniques omnipresent in AI research – and their merits and shortcomings for modelling aspects of design, as described by imaginative constructivism. To demonstrate and explain the kind of creative process expressed by the notion of Brouwer machine we contrasted it with a system using genetic algorithms for scientific law discovery.

*Dr. Akın Osman Kazakçı*
*Assistant Professor*
*Mines Paristech, CGS.*
*60, Bd. Saint-Michel, 7527, Paris Cedex 06*
*T: +33 1 40 51 93 12*
*F: +33 1 40 51 90 65*
*akin.kazakci@mines-paristech.fr*
*http://www.cgs-mines-paristech.fr/equipe/kazakci/*